# Assessing reliability of explanations in unbalanced datasets: a use-case on the occurrence of frost events

Ilaria Vascotto[1,*], Valentina Blasone[1], Alex Rodriguez[1,2], Alessandro Bonaita[3] and Luca Bortolussi[1]

[1]*Department of Mathematics, Informatics and Geosciences, University of Trieste, Trieste, Italy*
[2]*The Abdus Salam International Center for Theoretical Physics, Trieste, Italy*
[3]*Assicurazioni Generali Spa, Milan, Italy*

## Abstract

The usage of eXplainable Artificial Intelligence (XAI) methods has become essential in practical applications, given the increasing deployment of Artificial Intelligence (AI) models and the legislative requirements put forward in the latest years. A fundamental but often underestimated aspect of the explanations is their robustness, a key property that should be satisfied in order to *trust* the explanations. In this study, we provide some preliminary insights on evaluating the reliability of explanations in the specific case of unbalanced datasets, which are very frequent in high-risk use-cases, but at the same time considerably challenging for both AI models and XAI methods. We propose a simple evaluation focused on the minority class (i.e. the less frequent one) that leverages on-manifold generation of neighbours, explanation aggregation and a metric to test explanation consistency. We present a use-case based on a tabular dataset with numerical features focusing on the occurrence of frost events.

## 1. Introduction

Nowadays, Artificial Intelligence (AI) and Machine Learning (ML) models have become essential tools for practitioners tackling real and complex problems. These include high-risk applications such as in healthcare, climate, and fraud detection, which require highly reliable models due to the serious consequences that incorrect or biased predictions may have. However, many real-world datasets in these domains are intrinsically unbalanced as the critical events, e.g. rare diseases, natural catastrophes, or fraudulent transactions, occur less frequently than the normal conditions. Dataset unbalance introduces significant challenges for the AI and ML models, often leading to biased model predictions. In this context, gaining insights on what lies behind a model's prediction is particularly valuable for enhancing its trustworthiness. Despite this, ML models are usually very complex and are often treated by practitioners as *black boxes*.

Explainable Artificial Intelligence (XAI) methods aim at improving the transparency of ML models and offer a set of tools that can be used to *open the black box* either via local or global explanations. In practical applications, while efforts are made towards the development of highly accurate ML models, the explainability aspect may at times be overlooked. This is in part an undesirable consequence of the new legislative requirements that have been proposed for high-risk applications, both in the GDPR [1] and AI Act [2]. The transparency requirement can at times result in the incautious use of XAI techniques to satisfy the legislative needs on such use-cases. As an example, practitioners may apply frequently cited methodologies, such as LIME [3] and SHAP [4], without fully understanding their theoretical backgrounds and feature-wise requirements. Careless application of XAI approaches to high-risk scenarios may reflect into unintended, yet harmful, consequences. In this respect, the robustness of the explanations is often left unconsidered, whereas it is a decisive aspect to increase model trustworthiness and reliability of the provided explanation. The robustness of explanations (also referred to as stability) can be broadly defined as the ability of an explanation method (or explainer) to produce similar explanations for similar inputs.

When evaluating unbalanced datasets, assessing the reliability of explanations becomes even more important. To better understand the problem, we can consider a simple example. Assume that a model $f(\cdot)$ reaches an accuracy of 99% on the training dataset for a real-world problem. If the dataset is balanced or with a limited factor of unbalance, we can conclude that the model is behaving appropriately. If instead we assume to be working with a highly unbalanced dataset, e.g. $Pr(y = 0) = 0.99$ and $Pr(y = 1) = 0.01$, we cannot come to the same conclusion. In fact, a 99% accuracy could be easily obtained with a model that always predicts the class 0, which is obviously not the expected behaviour. This example sparks an important consideration on explanation reliability: if the majority class, that is the most frequently occurring one, is easily predicted by the model, the derived explanations cannot be deemed trustworthy. Moreover, considering that most of the times the training of ML models with unbalanced data is specifically adjusted to concentrate on the minority class, we cannot take for granted that the model has correctly learned the specific facets of the majority class, even if the accuracy is high.

In this study, we address what has been discussed so far by presenting a preliminary evaluation of *if* and *how* explanations can be trusted in the context of unbalanced datasets. For this purpose, we take inspiration from the framework recently introduced in [5], which was proposed for evaluating the trustworthiness of model explanations. We propose to focus on minority class explanations, as accurately predicting rare events is crucial in high-risk applications. As a use-case, we take a real scenario where a ML model is trained to predict the occurrence of extreme weather events, starting from atmospheric tabular data. Specifically, we consider the occurrence of frost events, which are inherently rare compared to normal weather conditions, leading to a highly unbalanced dataset where instances of the extreme phenomena are significantly lower, when compared to non-event cases.

## 2. Methodology

An overview of the proposed method is illustrated in Figure 1. Let $\mathcal{D} = (\mathbf{X}, \mathbf{y})$ be a dataset with $N$ points and $m$ features where $\mathbf{y}$ is the vector of class labels. Let $y = 0$ indicate the majority class and $y = 1$ the minority class, such that $Pr(y = 0) \gg Pr(y = 1)$. Let $\mathcal{D}$ be divided into training, validation and test sets and let $f(\cdot)$ be a neural network trained on the training set. Let $e$ be an explanation method and $e(\mathbf{x})$ the explanation associated to the data point $\mathbf{x}$ and model $f$. If $e$ is a feature importance method, the explanation will be also referred to as feature attribution.

Let $\mathbf{x}$ be a data point of interest and let $\tilde{\mathbf{x}}$ be a perturbation such that $dist(\mathbf{x}, \tilde{\mathbf{x}}) < \epsilon$ with $\epsilon > 0$ with $dist$ a distance metric, e.g. the Euclidean distance. A local neighbourhood $\mathcal{N}$ of point $\mathbf{x}$ is defined as the set of perturbed data points that are sufficiently close to the original data point and for which a model $f$ predicts the same class as the original point, $\hat{y}$.

### 2.1. Neighbourhood Generation

In the context of XAI, a local explanation aims at understanding the reasoning behind a model's prediction for a specific input vector, say $\mathbf{x}$. A key property of local explanations is their *robustness*, as introduced in Section 1. By definition, locality plays a central role in the evaluation of local explanations but it can lead to misleading results when the data manifold is not properly taken into account. A neighbourhood $\mathcal{N}$ should not only be made of datapoints which are close to the original one, but also be faithful to the observed data distribution: only in this scenario the evaluation can be trustworthy.

Leveraging the manifold hypothesis, we are able to construct neighbourhoods which are on-manifold as in [5]. Firstly, we apply the *k*-medoids clustering algorithm to the validation set, obtaining $k_{medoids}$ clusters that are, on average, of size $n_k = 10$. From each cluster $c$, we extract the medoid $\mathbf{x}^c$ as a representative, compute its $k_{nn} = 5$ nearest neighbours within the other ($k_{medoids} - 1$) cluster centres and randomly select from this set a medoid $\mathbf{x}^M$. For each datapoint in the test set, we predict the corresponding medoid cluster and retrieve its medoid's neighbours. If we assume that $\lambda$ represents the probability of perturbing a numerical variable $x_j$, then a perturbation $\tilde{\mathbf{x}}$ is computed feature-wise as:

$$\tilde{x}_j = (1 - \bar{\lambda}) \cdot x_j + \bar{\lambda} \cdot x_j^M \quad \text{with } \bar{\lambda} \leftarrow Beta(\lambda \cdot 100, (1 - \lambda) \cdot 100) \quad (1)$$

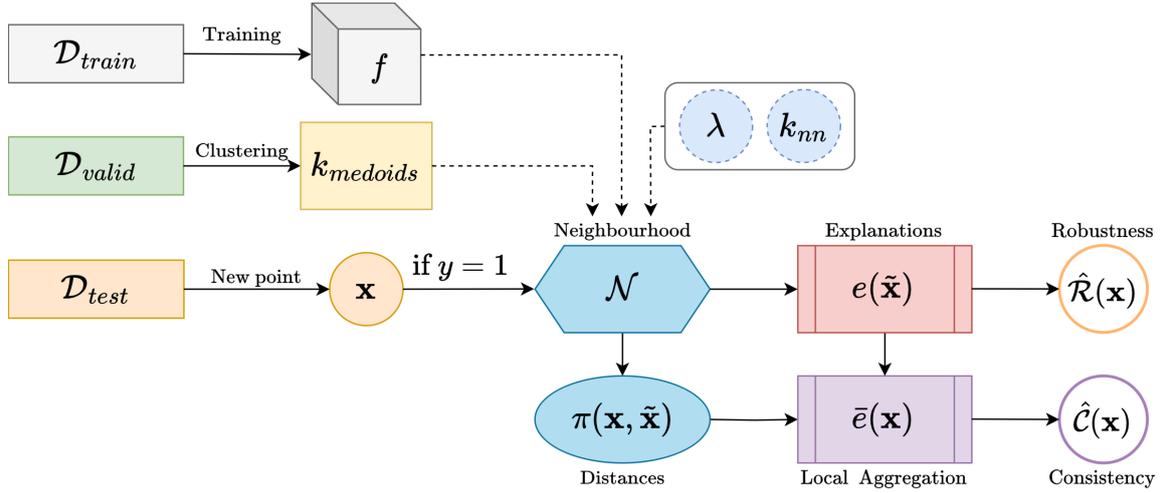

**Figure 1:** A visual representation of the proposed steps for minority class evaluation.

A neighbourhood $\mathcal{N}$ created following this perturbation scheme should be at least of size $n = 100$. A final filtering step is then performed to discard all perturbations that change the predicted class label.

## 2.2. Local Averaging

Given a neighbourhood of size $n$, costructed via Equation 1, we apply an explanation method $e$ to both the original data point and its perturbations, retrieving the local explanations $e(\mathbf{x})$ and $e(\tilde{\mathbf{x}})$, with $\tilde{\mathbf{x}} \in \mathcal{N}$.

For points belonging to the minority class $y = 1$, we can compute a local weighted attribution in order to retain local information. In particular, the averaged explanation is:

$$\bar{\mathbf{e}}(\mathbf{x}) = \frac{\sum_{\tilde{\mathbf{x}} \in \mathcal{N}} \mathbf{e}(\tilde{\mathbf{x}}) \cdot \pi(\mathbf{x}, \tilde{\mathbf{x}})}{\sum_{\tilde{\mathbf{x}} \in \mathcal{N}} \pi(\mathbf{x}, \tilde{\mathbf{x}})} \quad \text{where } \pi(\mathbf{x}, \tilde{\mathbf{x}}) = \frac{1}{dist(\mathbf{x}, \tilde{\mathbf{x}})} \quad (2)$$

Using the aggregated explanation allows us to take into account the results of a data augmentation on the minority class, which is ensured to be faithful as it lays on-manifold. Explanations are weighted according to the distance between the original data point and the perturbed one: same-class perturbations which are more distant will be given smaller weight in the aggregated explanation.

## 2.3. Proposed Evaluation

We assess the reliability of the minority class explanations by verifying if the locally weighted attribution is robust. In particular, we compare two locally-computed scores on $\mathcal{N}$.

The first score $\hat{\mathcal{R}}(\mathbf{x})$ - *local robustness* [5] - is computed locally by considering the Spearman rank correlation $\rho$ between the explanations of the point of interest and those of its perturbed neighbours.

$$\hat{\mathcal{R}}(\mathbf{x}) = \frac{1}{|\mathcal{N}|} \sum_{\tilde{\mathbf{x}} \in \mathcal{N}} \rho(e(\mathbf{x}), e(\tilde{\mathbf{x}})) \quad (3)$$

The second score $\hat{\mathcal{C}}(\mathbf{x})$, referred to as *consistency*, computes the rank correlation between the original explanation and the locally weighted averaged one (Equation 2).

$$\hat{\mathcal{C}}(\mathbf{x}) = \rho(e(\mathbf{x}), \bar{e}(\mathbf{x})) \quad (4)$$

The consistency score allows us to investigate if the aggregated explanation is locally faithful to the original one, ensuring that it summarizes useful information on the model decision making.

## 3. Use-case

To get some preliminary insight on *if* and *how* explanations on unbalanced datasets can be *trusted*, we apply the described methodology (Figure 1) to a real-world use-case. The case study was chosen to reflect what might be the needs of a practitioner, who finds himself working with an unbalanced dataset and having to apply XAI techniques, and thus wanting to determine whether the explanations received can be trusted or not. The code implementation is available on Github.[1]

### 3.1. Dataset Description

We selected the problem of identifying the occurrence of frost events, using atmospheric data as predictors. Frost is critical for agriculture, but rare and difficult to predict. A model that quantifies the occurrence of frost can be the first step for practitioners to estimate agricultural business policies in regions where data are limited. We used the the publicly available ERA5 reanalysis data [6] from the European Centre for Medium-Range Weather Forecasts (ECMWF) to construct the input features. We selected 8 numerical variables, each standardized to zero mean and unit variance. Since all features are numerical, we used the Euclidean distance to compute $\pi$ in Equation 2. Data are aggregated by days and by municipalities and only the spring/summer months are retained. Target data were instead obtained from proprietary datasets on insurance policies and claims, and take on a value of 0 or 1, whether or not the event occurred on that particular day/municipality. We considered a period of 15 years (2009-2024) over the territory of Poland. The obtained dataset is highly unbalanced with dataset class frequencies of 0.99 and 0.01 for the majority class 0 and minority class 1, respectively.

### 3.2. Model and Training

The model's architecture is a simple fully connected neural network, which takes the input features and processes them through five consecutive linear layers. The first four layers are followed by a ReLU activation function, while the final output layer applies a sigmoid activation to produce a probability score for classification. Data were split in three subsets: train (75%), validation (15%) and test (10%). The split was performed using a stratified approach based on municipalities, ensuring that all days of a municipality are assigned to the same set, either training or validation/test. We trained the model using the focal loss (FL), first introduced by [7] and shown in Equation 5. FL modifies the cross entropy (CE) loss formulation by introducing a hyper-parameter $\gamma$ which down-weights the contribution of easy examples and guides the model to focus on hard examples. Additionally, the hyper-parameter $\alpha$ is used to handle the class imbalance. We performed a random search to tune the loss hyper-parameters and found that $\gamma = 2.5$ and $\alpha = 0.75$ give the best performance on the validation set.

$$FL(p) = -\alpha(1-p)^\gamma \log(p) \tag{5}$$

We trained the model for 100 epochs with a batch size of 256, using the RAdam [8] optimizer with learning rate 0.0001. As metric to evaluate the model performance we used the F1-score, particularly suited for the case of unbalanced datasets. The results are reported in Table 1. We observe that the F1-score on the majority class 0 reaches its best value of 1.0 in all the three sets, while on the minority class 1 the obtained values are 0.66, 0.50 and 0.51 on the train, validation and test sets, respectively. Although the performance is not optimal, we consider the results achieved to be good given the imbalance ratio and especially the limited number of data points available for the minority class.

---

[1]https://github.com/ilariavascotto/Reliability_Unbalanced

**Table 1**
The results in terms of F1-score on the train, validation and test sets for the majority class 0 and the minority class 1 and the support, i.e. the number of points in each set.

|         | TRAIN    |          | VALIDATION |         | TEST     |          |
|---------|----------|----------|------------|---------|----------|----------|
|         | F1-score | Support  | F1-score   | Support | F1-score | Support  |
| Class 0 | 1.00     | ~ 225 000 | 1.00      | ~ 45 000 | 1.00    | ~ 30 000 |
| Class 1 | 0.66     | ~ 2 500  | 0.50       | ~ 450   | 0.51     | ~ 300    |

## 4. Preliminary Results

As model-agnostic methods such as LIME and SHAP have been proved to exhibit poor robustness, as shown in [9, 10], we focused on explanation methods which are specifically tailored to neural networks, as this is the model chosen in the use-case. In particular, we considered four explainability methods: Integrated gradients [11], DeepLIFT [12], Layerwise Relevance Propagation (LRP) [13] and the ensemble approach proposed in [5]. The first three methods are local post-hoc explanations that make use of the backpropagation procedure inherent in neural network's training to backpropagate a signal (often referred to as *relevance*) from the output layer to the input one. The methods differ in the used backpropagation rules and the possible presence of a baseline. Finally, the ensemble method is built from Integrated Gradients, DeepLIFT and LRP and aims at limiting the undesired effects of the disagreement problem [14], by providing a weighted average of multiple explanations.

We evaluated the data points belonging to the test set. We firstly investigated the impact of the neighbourhood generation on both the majority and minority class. In particular, we considered both a random neighbourhood generation, in which Gaussian noise $\epsilon \sim \mathcal{N}(0, \sigma^2)$ is added the numerical variables, and the medoid-based approach presented in Subsection 2.1. By comparing the robustness of Integrated Gradients (Figure 2) we show that there is indeed a difference between the two neighbourhood generating approaches and that the minority class is the most affected one. In fact, the random-generating scheme is associated with a larger distribution shift in the minority class than in the majority class. The negative effect of using a random neighbourhood is also evident if we consider the other explainability methods analysed. These results support our hypothesis that, in highly unbalanced datasets, the minority class is particularly vulnerable and requires careful analysis.

We then focused on the minority class data points ($y = 1$) and we built a neighbourhood of size $n = 100$ for each point, following the medoid-based generation scheme with hyperparameters $k_{nn} = 5$ and $\lambda = 0.05$. The hyperparameters were set through a random search to ensure that, on the validation set, at least 95% of the generated data points were classified as their corresponding original ones.

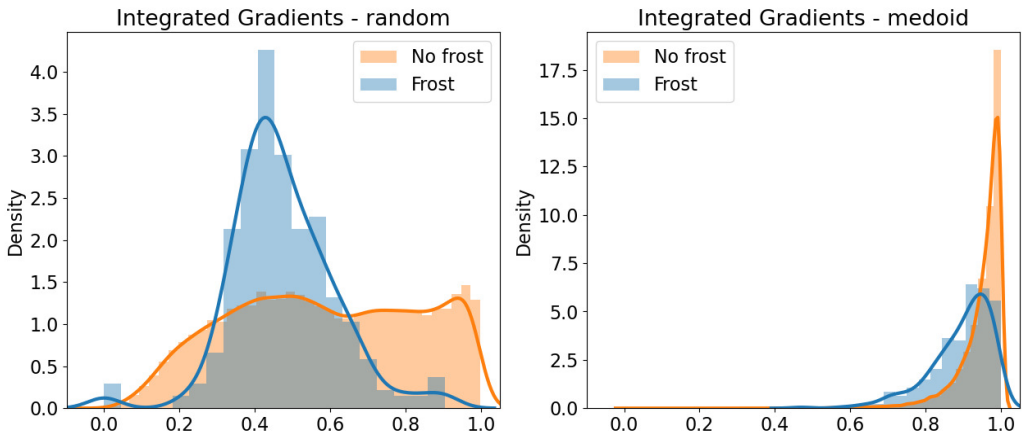

**Figure 2:** The robustness of the Integrated Gradients method computed with a random neighbourhood (left) and a medoid-based one (right). In blue, the distribution of the robustness score for the frost events (minority class) and in orange for the non-frost events (majority class).

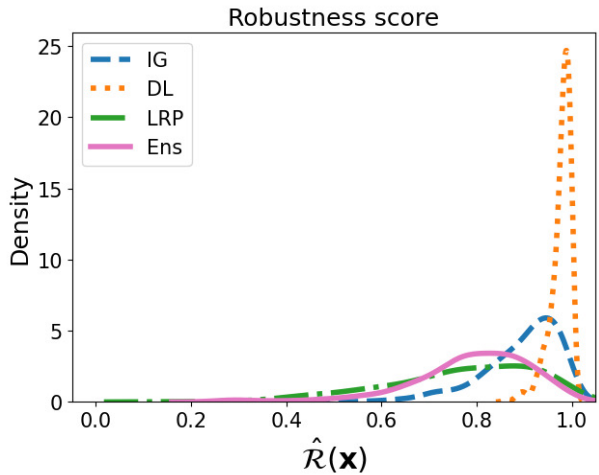

**Figure 3:** The robustness computed with a medoid-based neighbourhood for the four explainability methods: Integrated Gradients (IG), DeepLIFT (DL), LRP and Ensemble (Ens).

Figure 3 shows that the explanations may be more or less locally robust, depending on the considered method. We observe that the less stable method in this case-study is LRP, as confirmed by the lowest overall mean value and large standard deviation presented in Table 2. This technique at times suffers from the vanishing gradient problem, returning zero-vector attributions as a result. To limit the effect of uninformative explanations, we selected the epsilon propagation rule, as it was the one minimizing the vanishing gradient problem on this dataset. The choice of the propagation rule can significantly impact the results and should be carefully selected as a dataset-specific feature. When comparing the ensemble method robustness to the other two explainers, we need to take into account that its robustness is also influenced by LRP.

We showed that the individual explanations lie in a robust area of the data-manifold, as proven by the robustness scores derived on the test set for the four explainers (the larger the score, the more robust the area). However, this is not enough as we also need to ensure that the explanations are meaningful.

Our aim is to compute the locally weighted attribution $\bar{e}$ via Equation 2 and to test if it is consistent with the original point's explanation. This is done by computing the consistency score $\hat{\mathscr{C}}(\mathbf{x})$ via Equation 4. Since we showed that the data points lie in a robust area of the data space, we can leverage on-manifold information to enrich the set of data points predicted to be in class $y = 1$ and produce an individual explanation which takes into account also the neighbouring points. Increasing the amount of information considered when providing an individual explanation for a point of the minority class helps in understanding the decision making process of the model.

Figure 4 presents the consistency scores of the four explainers on the test set: we observe that the two methods which have higher robustness scores $\hat{\mathscr{R}}(\mathbf{x})$, namely Integrated Gradients and DeepLIFT, are associated to larger consistency scores $\hat{\mathscr{C}}(\mathbf{x})$. This was expected as the local explanations are more similar to one another and the averaged explanation remains highly correlated with the original one. In contrast, it is interesting to note that both LRP and the ensemble aggregation, despite having a lower robustness score due to the influence of the LRP method, are still able to produce satisfyingly consistent aggregated explanations within the generated neighbourhood.

Table 2 presents the mean robustness and consistency scores with the corresponding standard deviation reported within brackets. It can be seen that DeepLIFT is the most robust and most consistent method in the considered use-case, being associated with both the highest mean value and the lowest standard deviation. According to the practitioner's needs, the ensemble could also be deemed to be a valid candidate as it jointly considers the effects of multiple explanations, addressing possible inconsistencies by construction, and proposing satisfying robustness and consistency scores.

The consistency scores show that retaining local information from carefully crafted on-manifold neighbourhoods can be beneficial for the minority class explanations. In particular, the locally averaged explanation maintains more local information than the individual explanation on a given data point and it has been proved to encapsulate explanations of a robust area of the data manifold.

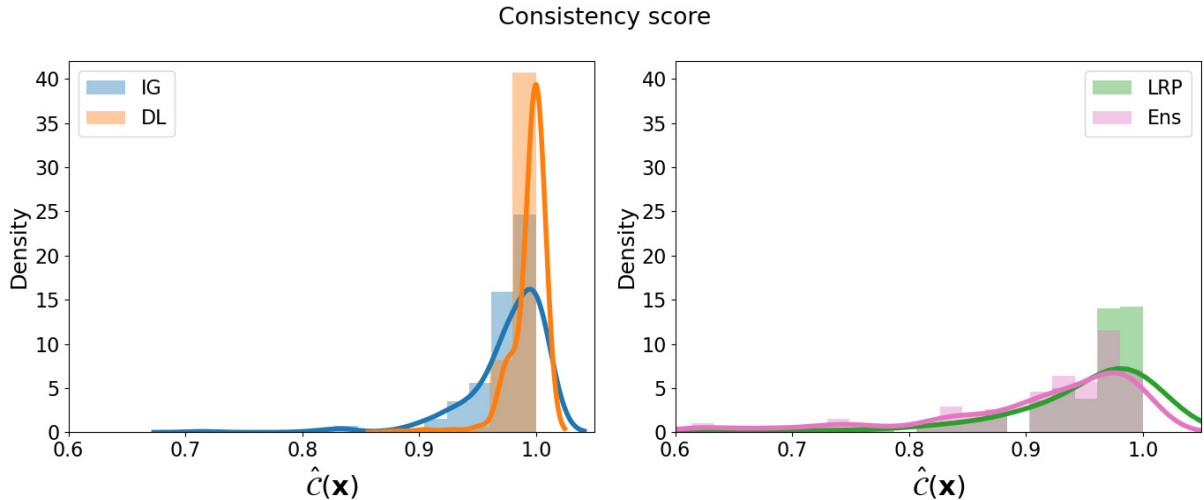

**Figure 4:** The consistency score of the four explainability methods: Integrated Gradients (IG) and DeepLIFT (DL) (left), LRP and Ensemble (Ens) (right).

**Table 2**
The average robustness and consistency scores on frost events for the four explainability methods. Values within brackets represent the standard deviation. Values in bold represent the maximum value over the four methods.

| Method | $\hat{\mathcal{R}}(\mathbf{x})$ | $\hat{\mathcal{C}}(\mathbf{x})$ |
|---|---|---|
| Integrated Gradients | 89.34 (±8.35) % | 97.56 (±3.58) % |
| DeepLIFT | **97.69 (±2.26)** % | **99.40 (±1.51)** % |
| LRP | 76.77 (±15.70) % | 89.86 (±19.95) % |
| Ensemble | 79.03 (±12.56) % | 89.20 (±13.73) % |

## 5. Conclusion and Future Developments

Despite being in its early stages of development, we argue that this line of research could bring to light useful insights in dealing with unbalanced datasets in practical use-cases. We believe that the analysis could be further enriched by defining a metric to quantify the quality of an explanation in such complex use-cases, considering both the robustness and the consistency as defined in Section 2. Moreover, it would be interesting to test how an uncertainty analysis on the minority class could be beneficial for the correct evaluation of these datasets from both a technical and a practitioner point of view.

As next steps, we plan to investigate how the results change as the performance of the neural network model varies and, at the same time, investigate the effects of a varying unbalancing ratio between the two classes. The analysis will be extended to different datasets, both public and private, in order to maintain the real-world aspect of interest to practitioners, but at the same time favour the reproducibility of the results. In future research stages, we also plan to investigate different machine learning models for improving performance with unbalanced dataset (e.g. by explicitly taking into account the spatio-temporal dimensions that often characterise many datasets) and adapt the choice of XAI methodologies and reliability reasoning accordingly.

## Acknowledgments

We wish to thank Assicurazioni Generali Spa for their support and interest in our work.

# Declaration on Generative AI

The authors have not employed any Generative AI tools.

# References

[1] European Commission, Regulation (EU) 2016/679 of the European Parliament and of the Council of 27 April 2016 on the protection of natural persons with regard to the processing of personal data and on the free movement of such data, and repealing Directive 95/46/EC (General Data Protection Regulation) (Text with EEA relevance), 2016. https://eur-lex.europa.eu/eli/reg/2016/679/oj.

[2] European Commission, Proposal for a Regulation of the European Parliament and of the Council laying down harmonised rules on artificial intelligence (Artificial Intelligence Act) and amending certain union legislative acts (COM(2021) 206 final), 2021. https://eur-lex.europa.eu/legal-content/EN/TXT/?uri=celex%3A52021PC0206.

[3] M. Ribeiro, S. Singh, C. Guestrin, "why should I trust you?": Explaining the predictions of any classifier, in: J. DeNero, M. Finlayson, S. Reddy (Eds.), Proceedings of the 2016 Conference of the North American Chapter of the Association for Computational Linguistics: Demonstrations, Association for Computational Linguistics, San Diego, California, 2016, pp. 97–101. URL: https://aclanthology.org/N16-3020/. doi:10.18653/v1/N16-3020.

[4] S. M. Lundberg, S.-I. Lee, A unified approach to interpreting model predictions, in: Proceedings of the 31st International Conference on Neural Information Processing Systems, volume 2017-December of *NIPS'17*, Curran Associates Inc., 2017, pp. 4766−−4775. https://dl.acm.org/doi/10.5555/3295222.3295230.

[5] I. Vascotto, A. Rodriguez, A. Bonaita, L. Bortolussi, When can you trust your explanations? a robustness analysis on feature importances, 2025. arXiv:2406.14349.

[6] H. Hersbach, B. Bell, P. Berrisford, S. Hirahara, A. Horányi, J. Muñoz-Sabater, J. Nicolas, C. Peubey, R. Radu, D. Schepers, et al., The era5 global reanalysis, Quarterly journal of the royal meteorological society 146 (2020) 1999–2049. doi:10.1002/qj.3803.

[7] T.-Y. Lin, P. Goyal, R. Girshick, K. He, P. Dollár, Focal loss for dense object detection, IEEE Transactions on Pattern Analysis and Machine Intelligence 42 (2020) 318–327. doi:10.1109/TPAMI.2018.2858826.

[8] L. Liu, H. Jiang, P. He, W. Chen, X. Liu, J. Gao, J. Han, On the variance of the adaptive learning rate and beyond, 2021. arXiv:1908.03265.

[9] D. Slack, S. Hilgard, E. Jia, S. Singh, H. Lakkaraju, Fooling lime and shap: Adversarial attacks on post hoc explanation methods, in: Proceedings of the AAAI/ACM Conference on AI, Ethics, and Society, AIES '20, Association for Computing Machinery, New York, NY, USA, 2020, p. 180–186. URL: https://doi.org/10.1145/3375627.3375830. doi:10.1145/3375627.3375830.

[10] A. Gosiewska, P. Biecek, Do not trust additive explanations, 2020. URL: https://doi.org/10.48550/arXiv.1903.11420. arXiv:1903.11420.

[11] M. Sundararajan, A. Taly, Q. Yan, Axiomatic attribution for deep networks, in: Proceedings of the 34th International Conference on Machine Learning - Volume 70, ICML'17, JMLR.org, 2017, p. 3319–3328. https://dl.acm.org/doi/10.5555/3305890.3306024.

[12] A. Shrikumar, P. Greenside, A. Kundaje, Learning important features through propagating activation differences, in: Proceedings of the 34th International Conference on Machine Learning - Volume 70, ICML'17, JMLR.org, 2017, p. 3145–3153. https://dl.acm.org/doi/10.5555/3305890.3306006.

[13] S. Bach, A. Binder, G. Montavon, F. Klauschen, K.-R. Müller, W. Samek, On pixel-wise explanations for non-linear classifier decisions by layer-wise relevance propagation, PLoS ONE 10(7) (2015). doi:10.1371/journal.pone.0130140.

[14] S. Krishna, T. Han, A. Gu, S. Wu, S. Jabbari, H. Lakkaraju, The disagreement problem in explainable machine learning: A practitioner's perspective, Transactions on Machine Learning Research (2024). doi:10.21203/rs.3.rs-2963888/v1.